\documentclass[letterpaper, 10pt, conference]{ieeeconf}  

\IEEEoverridecommandlockouts                              

\usepackage{cite}
\usepackage{amsmath}

\usepackage{anyfontsize}
\usepackage{graphicx}
\usepackage{subcaption}
\usepackage[table]{xcolor}
\usepackage{makecell}
\usepackage{array}
\usepackage{caption}
\usepackage{graphicx}
\usepackage{float} 
\usepackage{subcaption}
\usepackage{fixltx2e}
\usepackage{stfloats}
\usepackage{url}
\usepackage{subfiles}

\makeatletter
\let\NAT@parse\undefined
\makeatother
\usepackage[colorlinks=true, linkcolor=blue, citecolor=blue, urlcolor=blue]{hyperref}

\usepackage{cleveref}
\usepackage{graphicx}
\usepackage{amssymb}
\usepackage{booktabs}
\usepackage{mathtools}
\usepackage{multirow}
\usepackage{xcolor}
\usepackage{tablefootnote}
\usepackage{pifont}
\usepackage[linesnumbered,ruled]{algorithm2e}
\usepackage{algpseudocode}

\creflabelformat{equation}{#2\textup{#1}#3}
\Crefname{equation}{Eq.}{Eqs.}
\Crefname{figure}{Fig.}{Figs.}
\Crefname{tabular}{Tab.}{Tabs.}

\title{\LARGE \bf
V-Link: Recovering Lost Visual Representations in Action DiT \\ for Vision-Language-Action Models
}

\author{\textbf{Yehao Lu$^{1,4}$, Jiarui Yang$^{2,4}$, Yuning Su$^{3,4}$, Yufeng Xie$^{4}$, Yu Zhong$^{4}$, Yazhou Zhang$^{4}$, Haiyu Lan$^{4}$,} \\
\textbf{Kaixiang Lu$^{4}$, Peiwen Lin$^{4}$, Chuang Wang$^{4}$, Zequn Qin$^{1}$, Enyu Li$^{4\dagger}$, Xi Li$^{1\dagger}$}\\
$^1$Zhejiang University, $^2$The Hong Kong University of Science and Technology (Guangzhou) \\
$^3$Simon Fraser University, $^4$AGIBOT
\thanks{Work done as a research intern at AGIBOT.}%
\thanks{$\dagger$ Corresponding authors.}%
}

\begin{document}

\maketitle

\begin{abstract}
Vision-language-action (VLA) models provide a scalable path toward generalist robotic manipulation by integrating visual perception, language understanding, and continuous action control. However, we reveal a critical limitation of VLA architectures: the action expert has limited access to the 3D geometric and 2D semantic information available in VLM features. This accessibility gap weakens perceptual grounding and limits performance on fine-grained robotic manipulation. To address this issue, we propose V-Link, which explicitly recovers visual representations during the vision-language (VL) to action (A) feature transfer.  Specifically, V-Link learns complementary Spatial and Semantic Query representations within the VLM and injects them into Action DiT through asymmetric pathways. Semantic Queries complement the original VLM image tokens, whereas Spatial Queries provide dedicated geometric conditioning for spatially grounded action generation. Across LIBERO, LIBERO-Plus, and RoboTwin 2.0, our V-Link improves the average success rate over base model GR00T N1.6 by +1.9\%, +31.2\%, and +18.8\%, respectively. On the AGIBOT A3 Ultra, V-Link further achieves gains of +20\% and +24\% on two real-world humanoid tasks. 
\end{abstract}

\section{Introduction}
Vision-language-action (VLA) models have emerged as a foundational paradigm for robotic manipulation by mapping visual observations and language instructions into continuous action spaces. 
Among them, dual-system architectures couple a vision-language model (VLM) for rich scene understanding and task reasoning with a lightweight action expert for high-frequency robot control. 
Central to this design is vision-language (VL) to action (A) feature transfer, which determines what information from the VLM is available to the action expert.
In representative models such as NVIDIA GR00T N1.6 \cite{Gr00t_n1}, only the VLM’s final-layer features are transferred to the action expert.
\textbf{However, a key question remains: does this final-layer transfer provide the action expert with sufficient access to the 2D and 3D perceptual cues required for precise manipulation?}

\begin{figure}[!t] 
    \centering
    \includegraphics[width=0.45\textwidth]{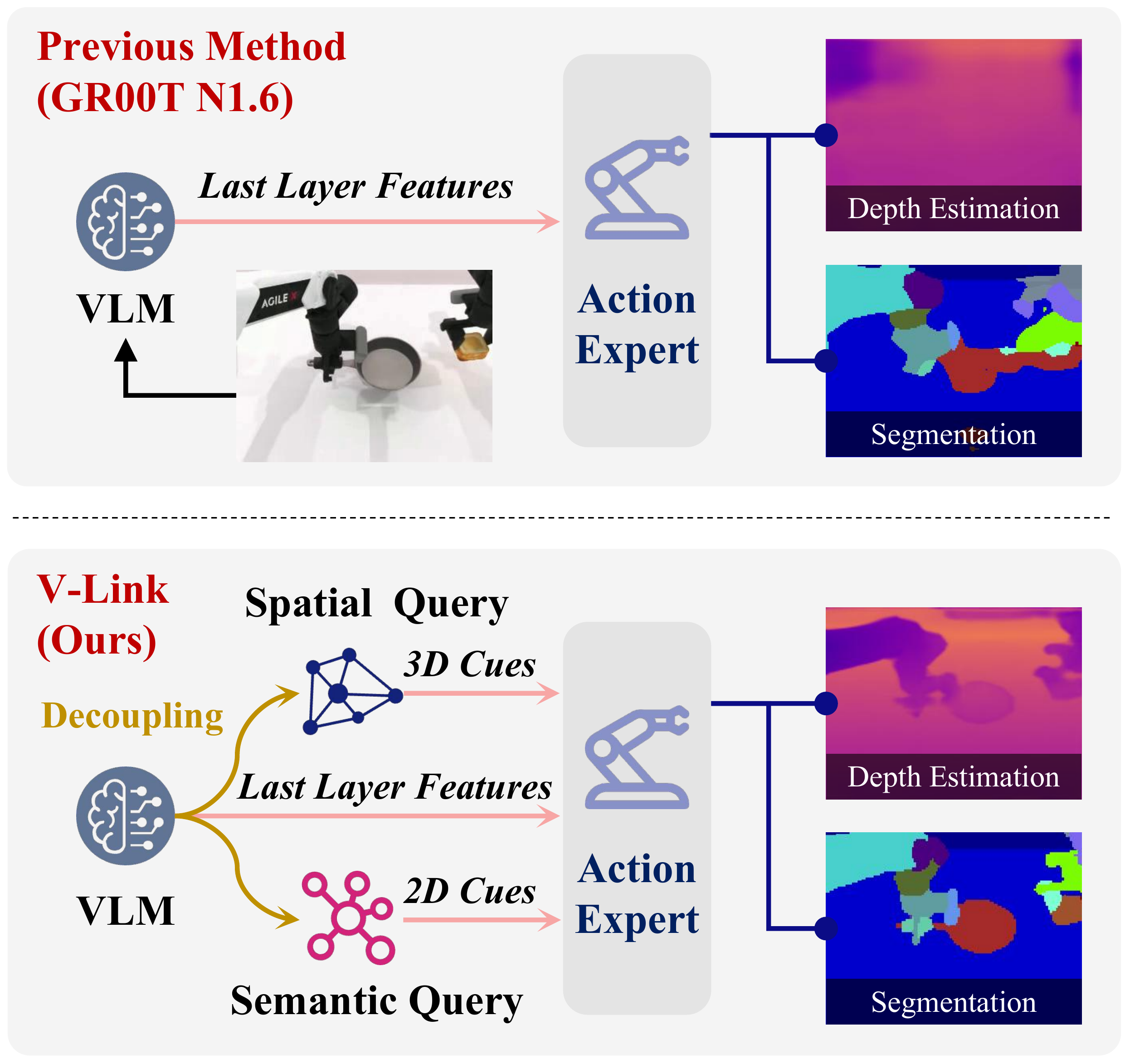}
    \caption{\textbf{Key insight.} GR00T N1.6 exhibits a visual representation accessibility bottleneck during VL-to-A feature transfer. Our V-Link addresses this bottleneck by learning complementary Spatial and Semantic Query representations and injecting them into Action DiT. The resulting Action DiT features support substantially better depth estimation and semantic segmentation, providing qualitative evidence of visual representation recovery.}
    \label{teaser}
    \vspace{-0.3cm}
\end{figure}

\begin{figure*}[!t] 
    \centering
    \includegraphics[width=0.95\textwidth]{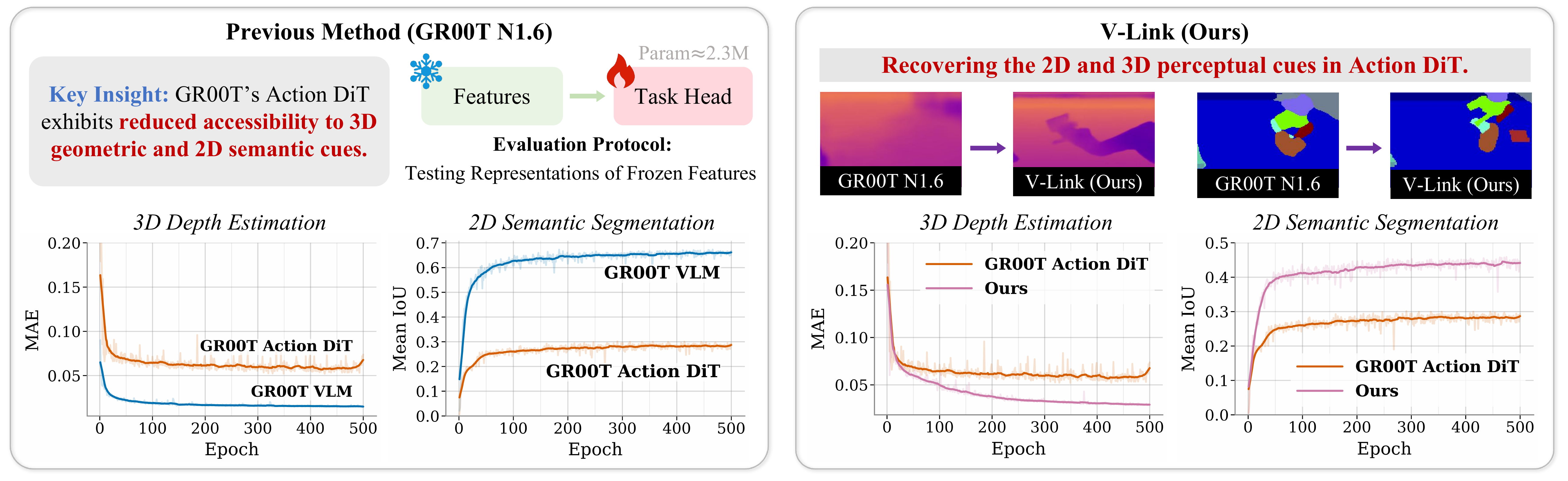}
    \caption{\textbf{Diagnostic evaluation protocol.} We freeze the VLM and Action DiT features and train only lightweight task heads for depth estimation and semantic segmentation. Test-set performance serves as a controlled measure of visual representation accessibility. \textbf{Left:} Compared with VLM features, GR00T N1.6 Action DiT features exhibit a pronounced accessibility gap. \textbf{Right:} V-Link substantially narrows this gap in Action DiT on both diagnostic tasks.}
    \label{motivation}
\end{figure*}

To answer this question, we employ a diagnostic protocol inspired by Spatial Forcing \cite{Spatial-forcing} to measure representation accessibility. Specifically, we freeze the GR00T N1.6 VLM and Action DiT features, attach lightweight depth-estimation and semantic-segmentation heads to these features, and optimize only the task heads under identical supervision.
With the underlying representations frozen, test-set performance serves as a controlled measure of the accessibility of 3D geometric and 2D semantic information.
As shown in \Cref{motivation} (left), the diagnostic heads achieve substantially better results with VLM features than with Action DiT features: depth MAE increases from 0.015 to 0.071, while segmentation mIoU decreases from 0.665 to 0.290. 
\textbf{These results reveal a pronounced visual representation accessibility gap across final-layer VL-to-A transfer.}

We hypothesize that the interaction between highly entangled final-layer VLM features and action-only supervision encourages Action DiT to exploit an optimization shortcut, prioritizing semantic cues that rapidly reduce the action loss while underutilizing structurally critical 3D spatial cues.
The qualitative results in \Cref{teaser} provide further support for this hypothesis: segmentation predictions decoded from GR00T Action DiT features remain coarse but recognizable, whereas the corresponding depth estimates recover little meaningful 3D structure.
Such asymmetric feature utilization can lead to a local optimum with insufficient 3D spatial grounding for precise robot manipulation.

Recent efforts to improve VLA perception, such as Spatial Forcing \cite{Spatial-forcing} and EVO-0 \cite{Evo-0}, primarily leverage representations from 3D foundation models to enhance spatial reasoning within the VLM. However, richer VLM representations do not guarantee that the action expert can access the same perceptual information after VL-to-A transfer.
Complementary work such as VLA-Adapter \cite{Vla-adapter} investigates efficient connections between the VLM and Action DiT, yet does not explicitly address the accessibility of transferred visual representations within the action expert.

Motivated by these observations, we propose V-Link, a representation-recovery framework that improves the accessibility of 3D geometric and 2D semantic information in Action DiT.
As illustrated in \Cref{teaser}, V-Link introduces learnable Spatial and Semantic Queries into the VLM to extract complementary perceptual representations. 
A tailored causal attention mask isolates the two query sets without disrupting the original multimodal computation, while training-only depth and segmentation heads specialize them for geometry and semantics, respectively.
V-Link then injects the learned query representations into Action DiT through asymmetric pathways. Semantic Queries complement the original VLM image tokens via parallel cross-attention, whereas Spatial Queries provide a dedicated geometry-conditioning pathway.
This design mitigates reliance on semantic shortcuts and makes both types of perceptual information directly accessible during action generation, as demonstrated by the representation diagnostics in \Cref{motivation} (right).

Extensive experiments across simulation benchmarks and real-world humanoid manipulation demonstrate consistent gains over GR00T N1.6. V-Link achieves average success-rate gains of +31.2\% on LIBERO-Plus, +18.8\% on RoboTwin 2.0, and +1.9\% on LIBERO. On the AGIBOT A3 Ultra, it achieves 98\% and 94\% success on autonomous power-on and power-off, outperforming GR00T N1.6 by +20\% and +24\%, respectively. These improvements incur only 1.58 ms of additional inference latency, from 43.21 ms to 44.79 ms, demonstrating substantial performance gains with minimal computational overhead.

Our contributions are summarized as follows:
\begin{itemize}
  \item We reveal a previously overlooked visual representation accessibility bottleneck in dual-system VLAs: final-layer VL-to-A transfer leaves the action expert with limited access to the 3D geometric and 2D semantic information required for precise manipulation.
  \item We propose V-Link, a representation-recovery framework that learns complementary Spatial and Semantic Query representations within the VLM and injects them into Action DiT through asymmetric pathways, improving access to both forms of perceptual information during action generation.
  \item Extensive experiments across three simulation benchmarks and two real-world humanoid tasks demonstrate consistent improvements over GR00T N1.6, while adding only 1.58 ms of inference latency.
\end{itemize}
\section{Related Work}

\subsection{VLA Architectures}
Existing VLA models can be broadly categorized into unified and dual-system architectures. Unified methods such as UniVLA \cite{UniVLA} and Cosmos Policy \cite{Cosmos_policy} model vision, language, and action within a shared transformer, while $\pi_{0.5}$ \cite{pi_05} and LingBot-VLA \cite{LingBot-VLA} use layer-wise attention to couple VLM and action representations. Dual-system designs pair a VLM with a separate action expert, decoupling multimodal reasoning from robot control. A representative model is GR00T N1.6 \cite{Gr00t_n1}, which transfers the VLM's final-layer features to Action DiT. Complementary studies improve efficiency through action tokenization \cite{pi0_Fast}, visual-token caching \cite{Vla-cache}, model scheduling \cite{Sp-vla}, or multi-frequency feature delivery \cite{UniFS}, while VLA-Adapter \cite{Vla-adapter} explores lightweight connections between the VLM and Action DiT. Prior work primarily focuses on architectural integration or execution efficiency. Instead, V-Link examines the accessibility of 3D geometric and 2D semantic representations within the action expert and explicitly recovers them during VL-to-A transfer.

\begin{figure*}[!t] 
    \centering
    \includegraphics[width=0.99\textwidth]{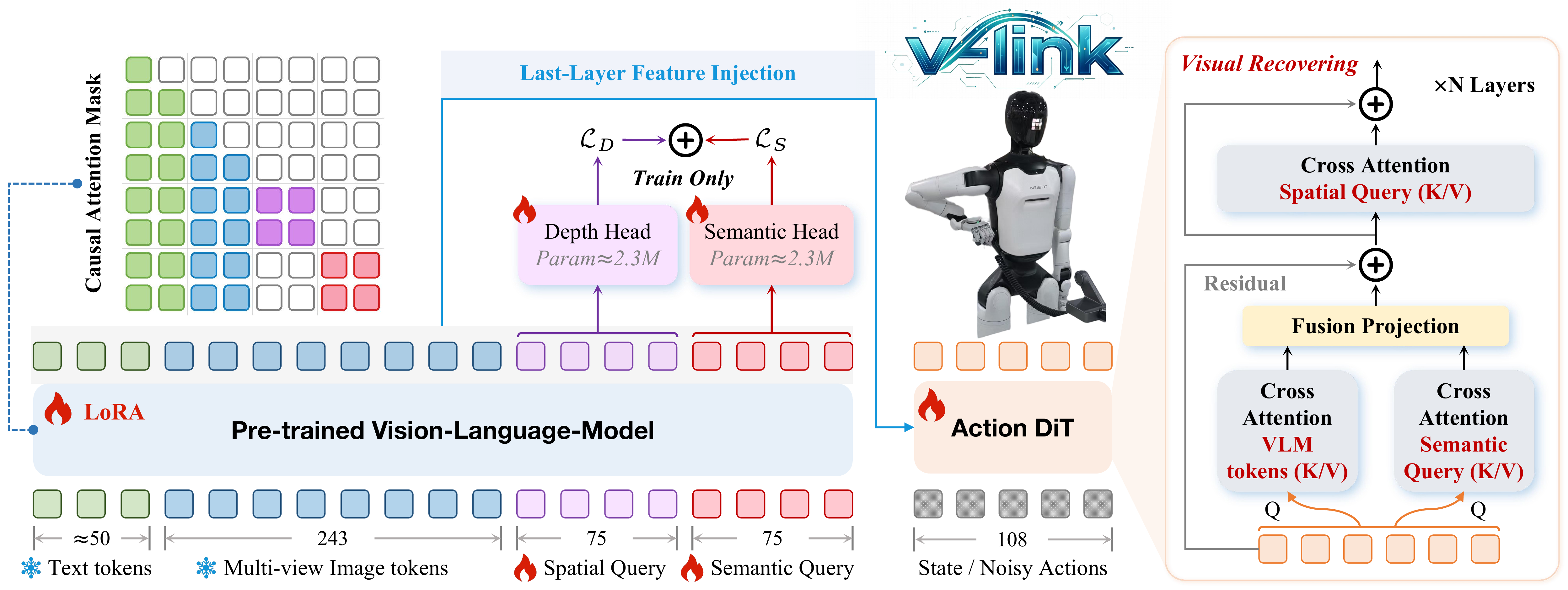}
    \caption{\textbf{The overall pipeline of V-Link.} V-Link introduces learnable Spatial and Semantic Queries into the pretrained VLM. A causal attention mask enables them to independently extract geometric and semantic cues from deep VLM features without disrupting the native multimodal token stream. Lightweight depth and semantic heads provide explicit supervision to disentangle these cues within the VLM latent space and are removed at inference. The resulting representations are injected into Action DiT through asymmetric pathways: Semantic Queries complement the VLM image tokens, whereas Spatial Queries form a mandatory geometry-conditioning branch that compels Action DiT to explicitly encode depth-aware representations rather than rely on semantic shortcuts.}
    \label{method_overview}
\end{figure*}

\subsection{Representation Enhancement}
Extensive research has focused on enhancing the visual perception of VLA. The vast majority of them prioritize improving the feature representation of the underlying VLM. For instance, GEAR-VLA \cite{GEAR-VLA} and SpatialVLA \cite{Spatialvla} incorporate trainable 3D spatial modules and align them with VLM representations. EVO-0 \cite{Evo-0} directly integrates 3D geometric features from VGGT \cite{Vggt} into the VLM. Spatial Forcing \cite{Spatial-forcing} further leverages VGGT features to distill VLM intermediate embeddings, eliminating the dependency on 3D foundation models at inference time. 
In parallel, a few works also aim to bolster the spatial reasoning of the action expert. PointVLA \cite{Pointvla} employs lightweight modules to inject 3D point cloud features. ABot-M0 \cite{Abot-m0} incorporates 3D features directly from 3D foundation models. 
In contrast to these approaches, our work aims to fundamentally optimize the VL to A feature transfer and mitigate the visual representation degradation during this process.

\section{Method}
\subsection{Overview}
The overall pipeline is illustrated in \Cref{method_overview}. We build upon GR00T N1.6 \cite{Gr00t_n1} as our base model. V-Link jointly learns complementary geometric and semantic representations within the VLM and injects them asymmetrically into Action DiT. Specifically, we introduce two sets of learnable tokens, termed Spatial Queries and Semantic Queries. A tailored causal attention mask isolates the two query sets while preserving the native image-text computation, allowing each set to aggregate relevant perceptual cues independently. Lightweight depth-estimation and semantic-segmentation heads provide training-only supervision, specializing the Spatial and Semantic Queries for geometry and semantics, respectively.

The learned query representations are then injected into Action DiT using an asymmetric attention design. The robot-state and noisy-action tokens serve as queries in two parallel cross-attention operations, with the VLM image tokens and Semantic Queries providing the respective keys and values. The resulting outputs are merged and added to the layer input through a residual connection, enriching the action features with complementary semantic context. The fused features subsequently undergo cross-attention with Spatial Queries, providing dedicated geometric conditioning for spatially grounded action generation. Following the alternating-attention design of GR00T N1.6, both query interactions are applied only to the image cross-attention layers.

\subsection{VLM Representations Decoupling}
When processing entangled VLM representations, Action DiT may preferentially exploit semantic cues over 3D geometry under action-only supervision. To counter this imbalance, we introduce learnable Spatial and Semantic Queries into the VLM input to separately capture geometric and semantic information:
\begin{equation}
    \mathbf E=[\mathbf E_t;\mathbf E_o;\mathbf E_d;\mathbf E_s],
\end{equation}
where $\mathbf E_t$ and $\mathbf E_o$ denote the language and multi-view image tokens, while $\mathbf E_d$ and $\mathbf E_s$ denote the Spatial and Semantic Query tokens, respectively.
Under the tailored causal attention mask $\mathbf M_q$, the two query sets independently attend to the VLM image and text tokens via self-attention. Across successive Transformer blocks, the query representations progressively aggregate rich visual cues:
\begin{equation}
[\mathbf E'_t, \mathbf E'_o, \mathbf E'_d, \mathbf E'_s]
= \mathcal R\bigl(\Phi_{\rm VLM}(\mathbf E;\mathbf M_q)\bigr),
\end{equation}
where $\Phi_{\rm VLM}$ denotes the VLM, and $\mathcal R$ denotes Qwen’s final RMSNorm layer. $\mathbf E'_t, \mathbf E'_o, \mathbf E'_d, \mathbf E'_s$ represent the final-layer VLM outputs corresponding to the image and text tokens, Spatial Query, and Semantic Query, respectively.

To explicitly encode depth information into the Spatial Query, we design a depth head $\Phi_{\rm Depth}$ to estimate dense depth maps $\hat{\mathbf D} \in \mathbb{R}^{H \times W \times V}$ for multiple camera views:
\begin{equation}
    \hat{\mathbf D}=\sigma_+(\Phi_{\rm Depth}(\mathbf E'_d)),
\end{equation}
where $\sigma_+(x)=\log(1+e^x)$ denotes the softplus function, which constrains the predicted metric depth to be positive. Our depth head comprises a projection layer $\mathcal P_d$, implemented as $\mathrm{LN}(2048)\rightarrow\mathrm{Linear}(2048,256)\rightarrow\mathrm{GELU}\rightarrow\mathrm{Dropout}$, two per-view transformer encoder layers $\mathcal T_d$, and a convolutional dense decoder $\mathcal C_d$ for depth prediction. 
With approximately 2.3M parameters, the depth head is intentionally lightweight, encouraging depth-relevant information to be captured by the Spatial Queries rather than the prediction head itself.
We supervise the predictions using ground-truth depth maps, with the depth loss defined as:
\begin{equation}
    \mathcal L_{depth}=\mathcal L_{\rm sm}+\lambda^d_1\mathcal L_{\rm rel}+\lambda^d_2\mathcal L_{\rm patch},
\end{equation}
where $\mathcal{L}_{\rm sm}$ denotes the smooth-$L_1$ loss for global absolute-scale depth supervision. $\mathcal{L}_{\rm rel}$ is the absolute-relative loss, which emphasizes near-field objects, robot arms, and manipulation regions while preventing distant backgrounds from dominating optimization. $\mathcal{L}_{\rm patch}$ is a top-$k$ hard-patch loss that mines locally difficult regions to avoid gradient dilution by abundant easy-background pixels. $\lambda_1^d=0.1$ and $\lambda_2^d=0.5$ are two constant coefficients.

The semantic head $\Phi_{\rm Seg}$ follows the same overall architecture as the depth head, differing only in its input queries, output dimensionality, and activation function:
\begin{equation}
    \hat{\mathbf Y}=\operatorname{Softmax}(\Phi_{\rm Seg}(\mathbf E'_s)),
\end{equation}
where $\hat{Y}\in\mathbb{R}^{H\times W\times C\times V}$ denotes the predicted 2D semantic segmentation logits, with $C$ and $V$ representing the numbers of semantic classes and camera views, respectively.
For the semantic head, we combine cross-entropy loss $\mathcal L_{CE}$ with soft dice loss $\mathcal L_{Dice}$. $\mathcal L_{CE}$ provides pixel-wise class supervision, whereas $\mathcal L_{Dice}$ directly optimizes region overlap and mitigates the bias induced by background dominance and class imbalance. The overall segmentation loss $\mathcal L_{seg}$ is formulated as:
\begin{equation}
    \mathcal L_{seg}=\mathcal L_{CE}+\lambda^s\mathcal L_{Dice}
\end{equation}
where $\lambda^s=0.5$ is a constant coefficient.

To stabilize optimization with randomly initialized Spatial and Semantic Queries, we first warm up the VLM LoRA parameters, both query sets, and their task heads using the same hyperparameters as subsequent joint training. After warm-up, all trainable components are jointly optimized. At inference, the auxiliary heads are removed, and no depth or segmentation annotations are required.



\begin{figure}[!t] 
    \centering
    \includegraphics[width=0.35\textwidth]{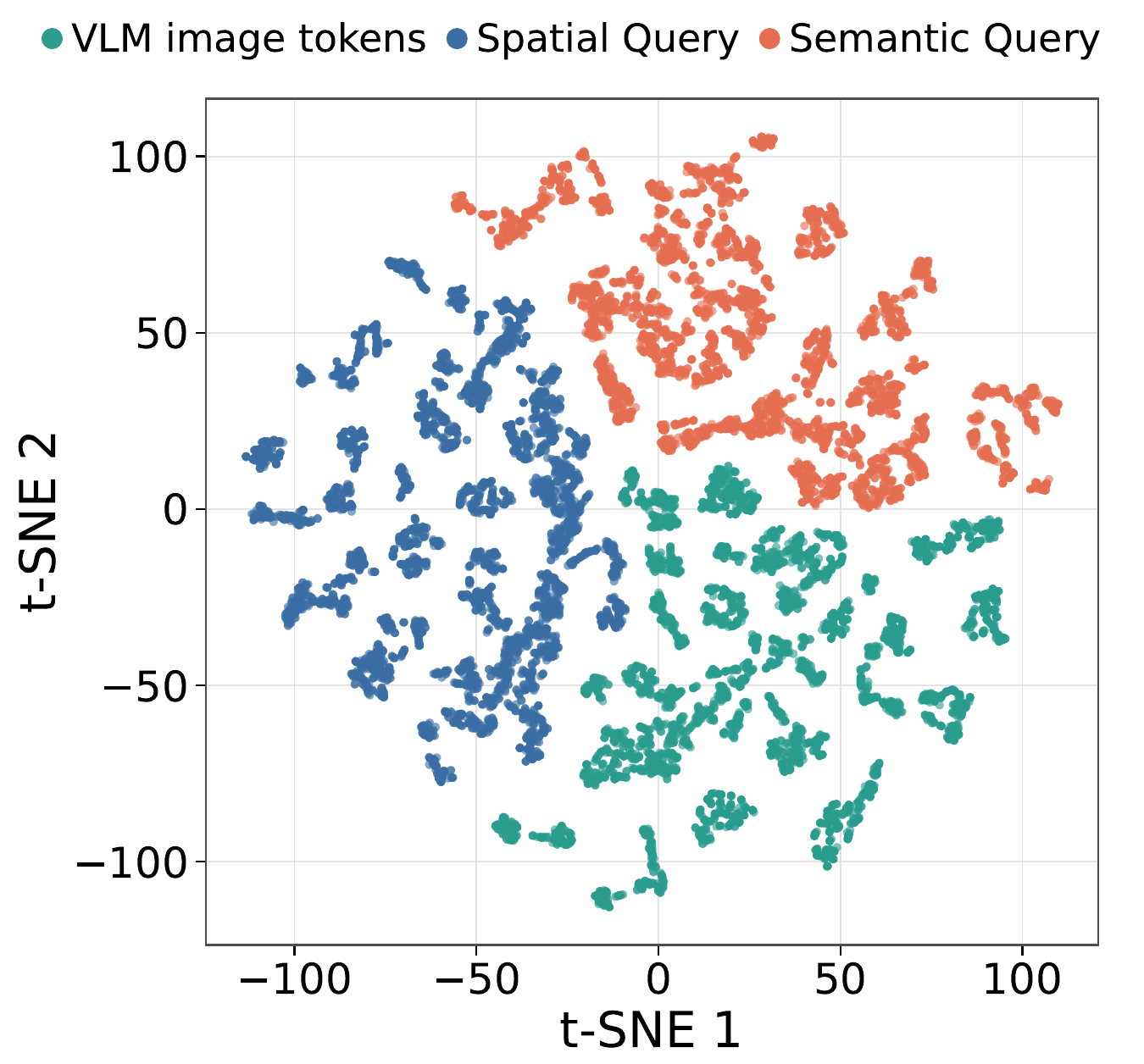}
    \caption{\textbf{t-SNE on RoboTwin2.0 rollouts.} We randomly sample 5,000 observations and extract the final-layer VLM outputs for image tokens, Spatial Queries, and Semantic Queries. Each token set is mean-pooled before projection. The separated distributions qualitatively indicate that the two query sets learn distinct representations.}
    \label{tsne}
\end{figure}

\subsection{Visual Representation Recovery in Action DiT}
GR00T N1.6 adopts a 32-layer AlternateVLDiT whose attention pattern comprises 8 repetitions of a 4-layer cycle:
\begin{equation}
\mathcal A_{\ell}=
\begin{cases}
\operatorname{CA}_{\mathrm{non\text{-}img}}, & \ell=4k,\\
\operatorname{SA}_{\mathrm{state/action}}, & \ell=4k+1,\\
\operatorname{CA}_{\mathrm{img}}, & \ell=4k+2,\\
\operatorname{SA}_{\mathrm{state/action}}, & \ell=4k+3,
\end{cases}
\quad k\in [0,7],
\end{equation}
where $\operatorname{CA}_{\mathrm{non\text{-}img}}$ and $\operatorname{CA}_{\mathrm{img}}$ denote cross-attention to valid non-image and image tokens, respectively, and $\operatorname{SA}_{\mathrm{state/action}}$ denotes self-attention over state and action tokens. 
Accordingly, the Spatial and Semantic Query representations are injected only at layers satisfying \(\ell=4k+2\). This confines V-Link to the image cross-attention layers while preserving the original alternating-attention schedule.

Our diagnostic analysis shows that the base model Action DiT retains coarse semantic cues but has limited access to 3D geometry. Motivated by this observation, we adopt an asymmetric injection design as illustrated in \Cref{method_overview}. Semantic Queries complement the original VLM image tokens through parallel cross-attention, whereas Spatial Queries subsequently provide dedicated geometric conditioning to the fused action features.
Formally, the robot state $\mathbf{s}$ and noisy action $\mathbf{x}_t$ are encoded into input tokens $\mathbf H_0$ for the Action DiT by $\operatorname{StateEnc}$ and $\operatorname{ActionEnc}$, respectively:
\begin{equation}
    \mathbf H_0=[\operatorname{StateEnc}(\mathbf s);\operatorname{ActionEnc}(\mathbf x_t,t)],
\end{equation}
At layer $l$, $\mathbf H_l$ serves as the query for two parallel cross-attention, with the VLM image tokens $\mathbf E'_o$ and Semantic Query $\mathbf E'_s$ independently providing the keys and values:
\begin{equation}
    \mathbf N_l=\operatorname{CA}_{\mathbf{VLM}}(\mathbf H_l,\mathbf E'_o),
\end{equation}
\begin{equation}
    \mathbf S_l=\operatorname{CA}_{\mathbf{Semantic}}(\mathbf H_l,\mathbf E'_s),
\end{equation}
The two cross-attention outputs are concatenated and projected by \(\mathrm{Fuse}\). A residual connection then adds the projected features to \(H_l\), yielding the semantically enriched representation \(Z_l\):
\begin{equation}
    \mathbf Z_l=\mathbf H_l+\operatorname{Fuse}(\mathbf N_l,g_l\mathbf S_l),
\end{equation}
where $g_l\in[0,1]$ is a learnable gating parameter. 
\(Z_l\) then serves as the query in a spatial cross-attention operation, with the Spatial Queries $\mathbf E'_d$ providing the keys and values. This operation explicitly conditions the action features on 3D geometry:
\begin{equation}
    \mathbf H_l'=\mathbf Z_l+\operatorname{CA}_{\mathbf{Spatial}}(\mathbf Z_l,\mathbf E'_d),
\end{equation}
Finally, the spatially conditioned feature \(H'_l\) is processed by an FFN to produce the output of the current layer $\mathbf H_{l+1}$:
\begin{equation}
    \mathbf H_{l+1}=\mathbf H_l'+\operatorname{FFN}(\mathbf H_l'),
\end{equation}

The overall training objective is defined as:
\begin{equation}
    \mathcal L=\mathcal L_{\mathrm{act}}+\lambda^a_1\mathcal L_{\mathrm{depth}}+\lambda^a_2\mathcal L_{\mathrm{seg}},
\end{equation}
where $\mathcal L_{\mathrm{act}}$ denotes the flow-matching loss for action generation, $\lambda^a_1=0.1$ and $\lambda^a_2=0.002$ are two constant coefficients.
\section{Experiments}
\label{sec:exp}
\subsection{Benchmarks and Evaluation Metrics}
We evaluate V-Link on 3 simulation benchmarks, including LIBERO \cite{Libero}, LIBERO-Plus \cite{Libero-plus}, and RoboTwin2.0 \cite{Robotwin-2.0}. For LIBERO and LIBERO-Plus, we follow the official protocols. All models are trained solely on standard LIBERO and evaluated on LIBERO-Plus without fine-tuning. For RoboTwin 2.0, we select 6 challenging tasks: Move Pillbottle Pad, Move Stapler Pad, Place Bread Skillet, Place Phone Stand, Press Stapler, and Turn Switch. These tasks span move, place, and contact interactions. We use 50 demonstrations per task, totaling 300 demonstrations for joint training. Ground-truth depth and semantic segmentation are used only during training and are not required at inference.
For real-world evaluation, we use the AGIBOT A3 Ultra and consider two tasks: Autonomous Power-On and Power-Off. We collect 100 teleoperated demonstrations per task. Success rates are reported over 50 trials per task on LIBERO and LIBERO-Plus, 100 trials per task on RoboTwin 2.0, and 50 trials per real-world task.


\begin{table}[!t]
    \centering
    \caption{\textbf{Comparison on LIBERO benchmark.}}
    \label{tab:LIBERO_comparison}
    \renewcommand{\arraystretch}{1.2}
    \resizebox{0.49 \textwidth}{!}{
    \begin{tabular}{lccccc}
        \toprule
        \textbf{Method} & \textbf{Spatial} & \textbf{Object} &
        \textbf{Goal} & \textbf{Long} & \textbf{Average} \\
        \hline

        \rowcolor{gray!15}
        \multicolumn{6}{c}{\textbf{2D VLA}} \\
        \hline

        OpenVLA \cite{Openvla}        & 84.7 & 88.4 & 79.2 & 53.7 & 76.5 \\
        Dita \cite{Dita}           & 84.2 & 96.3 & 85.4 & 63.8 & 82.4 \\
        CoT-VLA \cite{Cot-vla}        & 87.5 & 91.6 & 87.6 & 69.0 & 83.9 \\
        $\pi_0$-FAST \cite{pi0_Fast}    & 96.4 & 96.8 & 88.6 & 60.2 & 85.5 \\
        $\pi_0$ \cite{pi_0}       & 96.8 & 98.8 & 95.8 & 85.2 & 94.2 \\
        $\pi_{0.5}$ \cite{pi_05} &98.8 	&98.2 	&98.0 	&92.4 	&96.9 \\
        UniVLA \cite{UniVLA}         & 96.5 & 96.8 & 95.6 & 92.0 & 95.2 \\
        OpenVLA-OFT \cite{OpenVLA-OFT}    & 97.6 & 98.4 & 97.9 & 94.5 & 97.1 \\

        \hline
        \rowcolor{gray!15}
        \multicolumn{6}{c}{\textbf{3D VLA}} \\
        \hline

        SpatialVLA \cite{Spatialvla}         & 88.2 & 89.9 & 78.6 & 55.5 & 78.1 \\
        GeoVLA \cite{Geovla}            & 98.4 & 99.0 & 96.6 & 96.6 & 97.7 \\
        3D-CAVLA \cite{3d-cavla}           & 98.2 & 99.8 & 98.2 & 96.1 & 98.1 \\
        Spatial Forcing \cite{Spatial-forcing}    & 99.4 & 99.6 & 98.8 & 96.0 & 98.5 \\
        \midrule
        GR00T-N1.6 (Base)   & 99.3 & 99.2 & 98.4 & 92.9 & 97.4 \\
        \rowcolor{blue!10} V-Link (Ours)       & \textbf{99.5} 	& \textbf{100.0}  & \textbf{99.8} & \textbf{98.0} & \textbf{99.3}    \\
        \rowcolor{blue!10} $\Delta$ Ours - Base & \textcolor{red}{\textbf{+0.2}} & \textcolor{red}{\textbf{+0.8}}	& \textcolor{red}{\textbf{+1.4}} & \textcolor{red}{\textbf{+5.1}} & \textcolor{red}{\textbf{+1.9}}  \\
        \bottomrule
    \end{tabular}
    }
\end{table}

\begin{table*}[!t]
    \centering
    \caption{\textbf{Comparison on LIBERO-Plus benchmark.}}
    \label{tab:LIBERO-Plus-performance}
    \renewcommand{\arraystretch}{1.2}
    \resizebox{0.8 \textwidth}{!}{
    \begin{tabular}{lcccccccc}
        \toprule
        \textbf{Method} & \textbf{Camera} & \textbf{Robot} & \textbf{Language}
        & \textbf{Light} & \textbf{Background} & \textbf{Noise}
        & \textbf{Layout} & \textbf{Total} \\
        \midrule
        OpenVLA-OFT \cite{OpenVLA-OFT} & 56.4 & 31.9 & 79.5 & 88.7 & 93.3 & 75.8 & 74.2 & 69.6 \\
        $\pi_0$ \cite{pi_0}    & 13.8 & 6.0  & 58.8 & 85.0 & 81.4 & 79.0 & 68.9 & 53.6 \\
        $\pi_0$-Fast \cite{pi0_Fast}    & \textbf{65.1} & 21.6 & 61.0 & 73.2 & 73.2 & 74.4 & 68.8 & 61.6 \\
        Evo-Depth \cite{Evo-depth}  & 47.2 & 49.2 & 78.9 & 88.1 & 76.4 & 77.6 & 69.6 & 69.6 \\
        \midrule
        GR00T-N1.6 (Base)  & 21.2 & 40.3 & 35.5 & 65.1 & 76.7 & 27.4 & 51.0 & 43.8 \\
        \rowcolor{blue!10} V-Link (Ours) & 42.8 & \textbf{59.1} & \textbf{82.8} & \textbf{95.7} & \textbf{94.9} & \textbf{87.8} & \textbf{74.4} & \textbf{75.0}  \\
        \rowcolor{blue!10} $\Delta$ Ours - Base & \textcolor{red}{\textbf{+21.6}} & \textcolor{red}{\textbf{+18.8}} & \textcolor{red}{\textbf{+47.3}} & \textcolor{red}{\textbf{+30.6}} & \textcolor{red}{\textbf{+18.2}} & \textcolor{red}{\textbf{+60.4}} & \textcolor{red}{\textbf{+23.4}} & \textcolor{red}{\textbf{+31.2}} \\
        \bottomrule
    \end{tabular}
    }
\end{table*}


\begin{table*}[!t]
    \centering
    \caption{\textbf{Comparison on Robotwin2.0 benchmark.}}
    \label{tab:Robotwin_performance}
    \renewcommand{\arraystretch}{1.2}
    \resizebox{0.9 \textwidth}{!}{
    \begin{tabular}{lccccccc}
        \toprule
        \multirow{2}{*}{\textbf{Method}}
        & \multicolumn{2}{c}{\textbf{Move}}
        & \multicolumn{2}{c}{\textbf{Place}}
        & \multicolumn{2}{c}{\textbf{Contact}}
        & \multirow{2}{*}{\textbf{Average}} \\
        \cmidrule(lr){2-3}
        \cmidrule(lr){4-5}
        \cmidrule(lr){6-7}
        & \textbf{Pillbottle} & \textbf{Stapler}
        & \textbf{Bread Skillet} & \textbf{Phone Stand}
        & \textbf{Press Stapler} & \textbf{Turn Switch}
        & \\
        \midrule

        $\pi_0$ \cite{pi_0}
        & 21 & 0
        & 23 & 35
        & 62 & 27
        & 28.0 \\


        Xiaomi Robotics-0 \cite{Xiaomi-robotics-0}
        & 65 & 16
        & 63 & 44
        & 79 & 38
        & 50.8 \\


        EventVLA \cite{EventVLA}
        & 60 & 14
        & 64 & 64
        & 80 & 24
        & 51.0 \\

        $\pi_{0.5}$ \cite{pi_05}
        & 46 & 20
        & \textbf{76} & \textbf{65}
        & 79 & \textbf{52}
        & 56.3 \\

        \midrule

        GR00T N1.6 (Base)
        & 58 & 7
        & 23 & 54
        & 80 & 6
        & 38.0 \\

        \rowcolor{blue!10}V-Link (Ours)
        & \textbf{74}   & \textbf{22} 
        & 55   & 60
        & \textbf{97}   & 33
        & \textbf{56.8}    \\
        \rowcolor{blue!10} $\Delta$ Ours - Base
        & \textcolor{red}{\textbf{+16}}   & \textcolor{red}{\textbf{+15}}
        & \textcolor{red}{\textbf{+32}}   & \textcolor{red}{\textbf{+6}}
        & \textcolor{red}{\textbf{+17}}   & \textcolor{red}{\textbf{+27}}  
        & \textcolor{red}{\textbf{+18.8}}    \\
        \bottomrule
    \end{tabular}
    }
\end{table*}

\subsection{Implementation Details}
All models are initialized from GR00T N1.6 3B and trained in BF16 on 4 NVIDIA H100 GPUs using DeepSpeed ZeRO-2. We freeze the pretrained language and vision backbones and adapt the VLM using rank-128 LoRA. We use AdamW with an initial learning rate of \(1\times10^{-4}\), a weight decay of \(1\times10^{-5}\), cosine decay, and 5\% warm-up.
On LIBERO, models are trained for 80K steps with a global batch size of 160 using external and wrist camera views. The Spatial and Semantic Query sets each contain \(2\times5\times5\) tokens. The policy predicts 16-step action chunks.
On RoboTwin 2.0, models are trained for 120K steps with a global batch size of 128. Each query set contains \(3\times5\times5\) tokens. The policy predicts 15-step dual-arm action chunks.

\subsection{Comparisons with State-of-the-art Methods}
\Cref{tab:LIBERO_comparison} compares V-Link with state-of-the-art 2D and 3D VLAs on LIBERO. V-Link ranks first across all 4 suites, achieving an average success rate of 99.3\%. It outperforms GR00T N1.6 and Spatial Forcing by +1.9\% and +0.8\%, respectively, with the largest gain on LIBERO-Long (+5.1\%).


\Cref{tab:LIBERO-Plus-performance} evaluates robustness under 7 distribution shifts on LIBERO-Plus. V-Link achieves the highest overall success rate of 75.0\%, surpassing the previous best method by +5.4\% and GR00T N1.6 by +31.2\%. It ranks first in six settings, with particularly large gains over GR00T N1.6 under noise (+60.4\%), language (+47.3\%), and lighting (+30.6\%).

As shown in \Cref{tab:Robotwin_performance}, V-Link achieves the highest average success rate of 56.8\% across 6 RoboTwin 2.0 tasks, outperforming GR00T N1.6 by +18.8\%. Improvements are consistent across all tasks. These results demonstrate that recovering 3D spatial and 2D semantic representations substantially strengthens fine-grained manipulation.

\subsection{Ablation Study}


\begin{table}[t]
    \centering
    \caption{\textbf{Ablation study of auxiliary queries.}}
    \label{tab:query_ablation}
    \renewcommand{\arraystretch}{1.2}
    \resizebox{0.45 \textwidth}{!}{
    \begin{tabular}{lcccc}
        \toprule
        \textbf{Method} & \textbf{Move} & \textbf{Place}
        & \textbf{Contact} & \textbf{Average} \\
        \midrule
        GR00T N1.6      & 32.5 & 38.5 & 43.0 & 38.0 \\
        + Spatial Query & 44.5  & 54.0  & 55.0 &  51.2    \\
        + Semantic Query& 35.5     & 47.0     &  61.0    & 47.8     \\
        \rowcolor{blue!10}+ Both Query    &  \textbf{48.0}    & \textbf{57.5}     & \textbf{65.0}     & \textbf{56.8}     \\
        \bottomrule
    \end{tabular}
    }
\end{table}



\noindent \textbf{Q1: How do Spatial and Semantic Queries contribute to policy performance?}
\Cref{tab:query_ablation} shows that Spatial Queries improve the average success rate from 38.0\% to 51.2\% (+13.2\%), with substantial gains on Move (+12.0\%) and Place (+15.5\%), supporting their role in geometric grounding. Semantic Queries achieve 47.8\% (+9.8\%), with the largest gain on Contact (+18.0\%), suggesting improved localization of task-relevant interaction regions. Combining both yields the best performance of 56.8\% (+18.8\%), confirming their complementary roles. 


\noindent \textbf{Q2: How do task supervision and query injection contribute to policy performance?} 
\Cref{tab:supervision_ablation} and \Cref{tab:injection_ablation} isolate their effects. Query injection without task supervision improves the average success rate only from 38.0\% to 40.3\%, while task supervision without injection yields only 39.8\%. Combining both reaches 56.8\%, outperforming the two ablations by +16.5\% and +17.0\%, respectively. These results demonstrate their complementary roles: task supervision specializes the queries for 3D geometry and 2D semantics, while query injection makes these representations accessible to Action DiT.

\begin{figure}[!t] 
    \centering
    \includegraphics[width=0.42\textwidth]{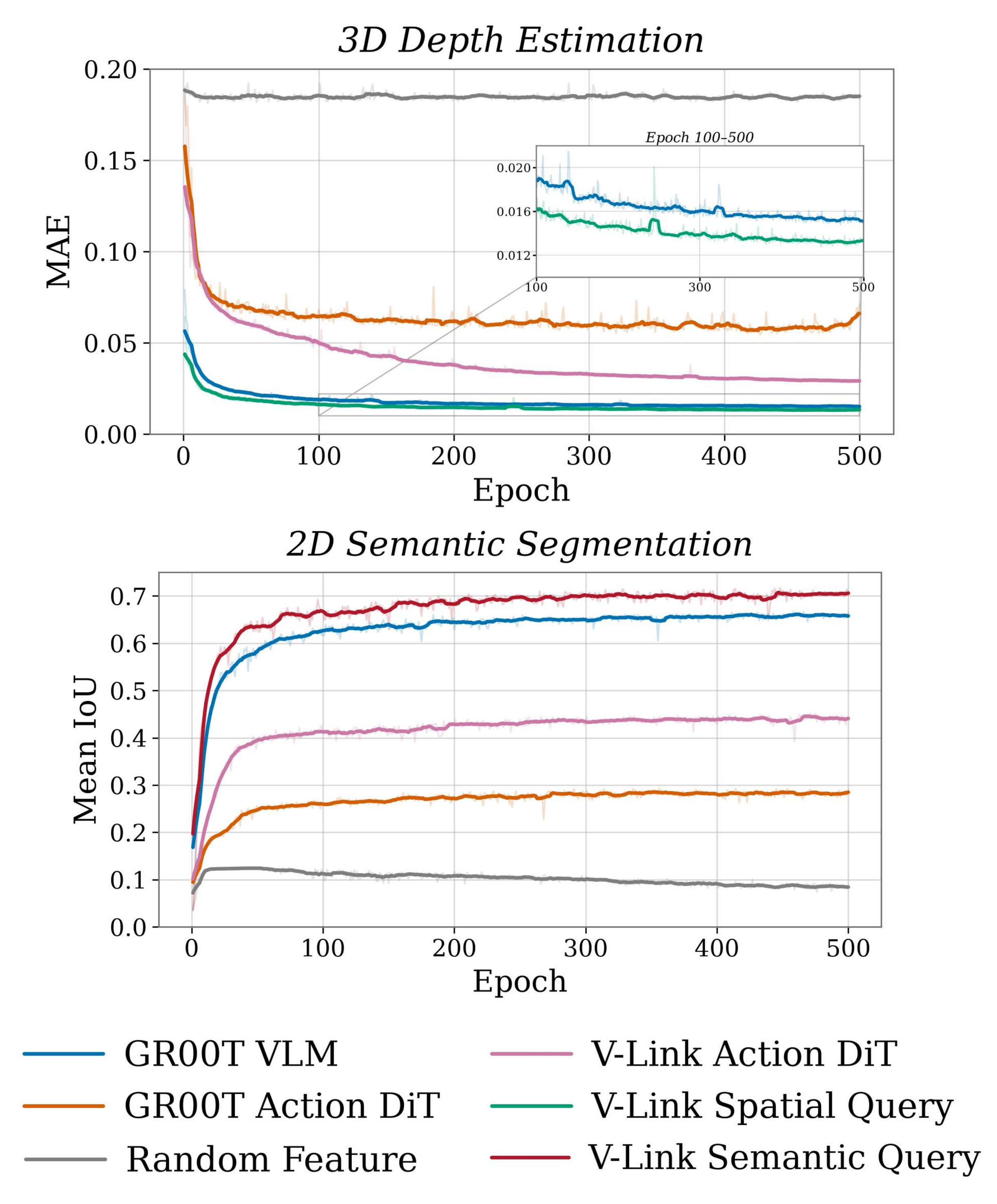}
    \caption{\textbf{Diagnostic evaluation of frozen visual representations.} 
    The poor results with random features show that diagnostic performance depends on the information encoded in the frozen features, rather than the task heads themselves. Spatial and Semantic Queries achieve the best depth-estimation and semantic-segmentation results, respectively, demonstrating task-specific specialization. V-Link Action DiT outperforms GR00T Action DiT on both tasks, confirming successful representation recovery.}
    \label{test_metric_curves_comparison1}
\end{figure}

\noindent \textbf{Q3: Does V-Link learn specialized visual representations and recover them in Action DiT?} Following our diagnostic protocol, we freeze each representation and train identical lightweight heads for depth estimation and semantic segmentation. Poor performance with random features shows that the results reflect representation accessibility rather than task-head capacity. As shown in \Cref{test_metric_curves_comparison1}, Spatial Queries achieve the lowest depth MAE, while Semantic Queries attain the highest segmentation mIoU, indicating their respective specialization. Moreover, V-Link Action DiT outperforms GR00T Action DiT on both tasks, demonstrating successful representation recovery after query injection.

\begin{table}[t]
    \centering
    \caption{\textbf{Ablation study of task supervision.} ``w/o Task Head" retains the Spatial and Semantic Queries and their injection mechanisms, while removing the depth and semantic heads and the ground-truth supervision.}
    \label{tab:supervision_ablation}
    \renewcommand{\arraystretch}{1.2}
    \resizebox{0.45 \textwidth}{!}{
    \begin{tabular}{lcccc}
        \toprule
        \textbf{Method} & \textbf{Move} & \textbf{Place}
        & \textbf{Contact} & \textbf{Average} \\
        \midrule
        GR00T N1.6      & 32.5 & 38.5 & 43.0 & 38.0 \\
        w/o Task Head & 35.5  & 38.5 & 47.0 & 40.3 \\
        \rowcolor{blue!10}w/ Task Head &  \textbf{48.0}  & \textbf{57.5}   & \textbf{65.0}  & \textbf{56.8}    \\
        \bottomrule
    \end{tabular}
    }
\end{table}

\begin{table}[t]
    \centering
    \caption{\textbf{Ablation study of auxiliary query injection.} ``w/o Injection" retains the Spatial and Semantic Queries, the depth and semantic heads, and the ground-truth supervision, but does not inject both queries into Action DiT.}
    \label{tab:injection_ablation}
    \renewcommand{\arraystretch}{1.2}
    \resizebox{0.45 \textwidth}{!}{
    \begin{tabular}{lcccc}
        \toprule
        \textbf{Method} & \textbf{Move} & \textbf{Place}
        & \textbf{Contact} & \textbf{Average} \\
        \midrule
        GR00T N1.6      & 32.5 & 38.5 & 43.0 & 38.0 \\
        w/o Injection & 40.5 & 28.5 & 50.5 & 39.8 \\
        \rowcolor{blue!10}w/ Injection &  \textbf{48.0} & \textbf{57.5} & \textbf{65.0} & \textbf{56.8}    \\
        \bottomrule
    \end{tabular}
    }
\end{table}

\begin{figure}[!t] 
    \centering
    \includegraphics[width=0.42\textwidth]{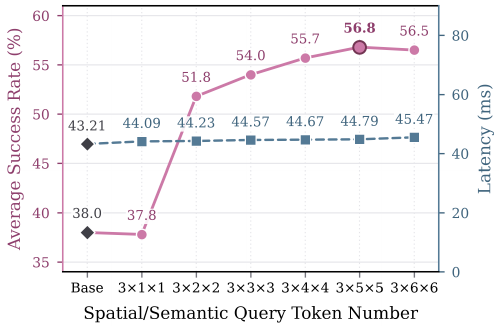}
    \caption{\textbf{Effect of Spatial/Semantic Query token number on performance and inference latency.} Latency is tested on a single H100 and averaged over 1,000 randomly sampled inputs with batch size 1, using torch.compile, 4 denoising steps, and 3 views input.}
    \label{query_token_ablation}
\end{figure}

\noindent \textbf{Q4: How does the number of Spatial/Semantic Query tokens affect policy performance and inference latency?}
As shown in \Cref{query_token_ablation}, performance improves as the query grid grows, but with diminishing returns. Increasing it from \(3\times1\times1\) to \(3\times2\times2\) raises the average success rate from 37.8\% to 51.8\%, indicating that sufficient query capacity is essential for extracting spatial and semantic information. With a \(3\times5\times5\) query grid, V-Link achieves the highest success rate of 56.8\%, outperforming GR00T N1.6 by +18.8\% with only 1.58 ms of additional inference latency. Further increasing the grid to \(3\times6\times6\) reduces performance slightly to 56.5\% and incurs additional latency. Therefore, \(3\times5\times5\) provides the best accuracy–efficiency trade-off.

\begin{figure}[!t] 
    \centering
    \includegraphics[width=0.48\textwidth]{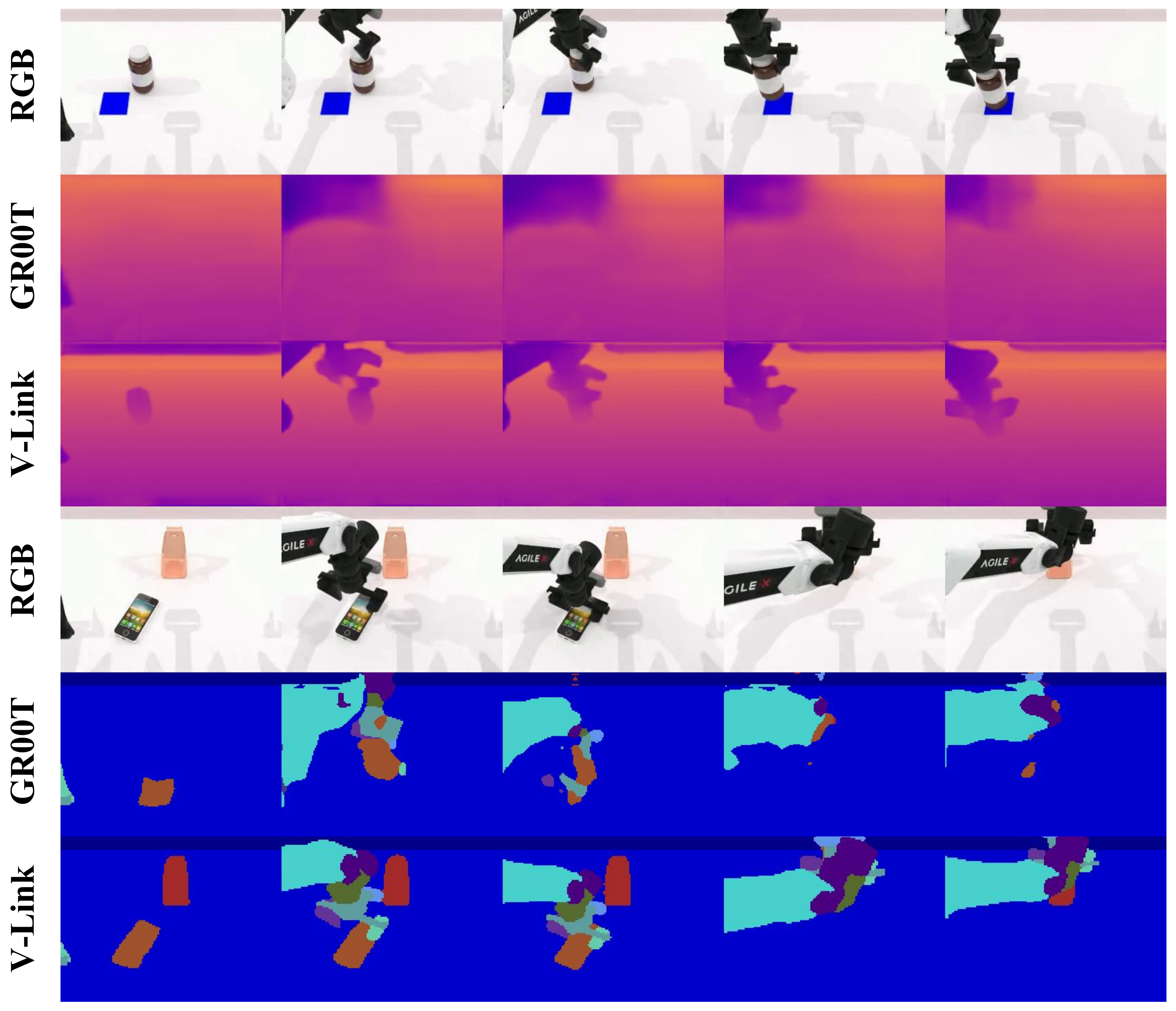}
    \caption{\textbf{Qualitative comparison.} The Action DiT of GR00T N1.6 exhibits severe visual representation degradation, retaining little 3D depth information and incomplete 2D semantics. V-Link effectively recovers the spatial and semantic representations within Action DiT.}
    \label{visualization}
    \vspace{-0.2cm}
\end{figure}

\begin{figure*}[!t] 
    \centering
    \includegraphics[width=0.9\textwidth]{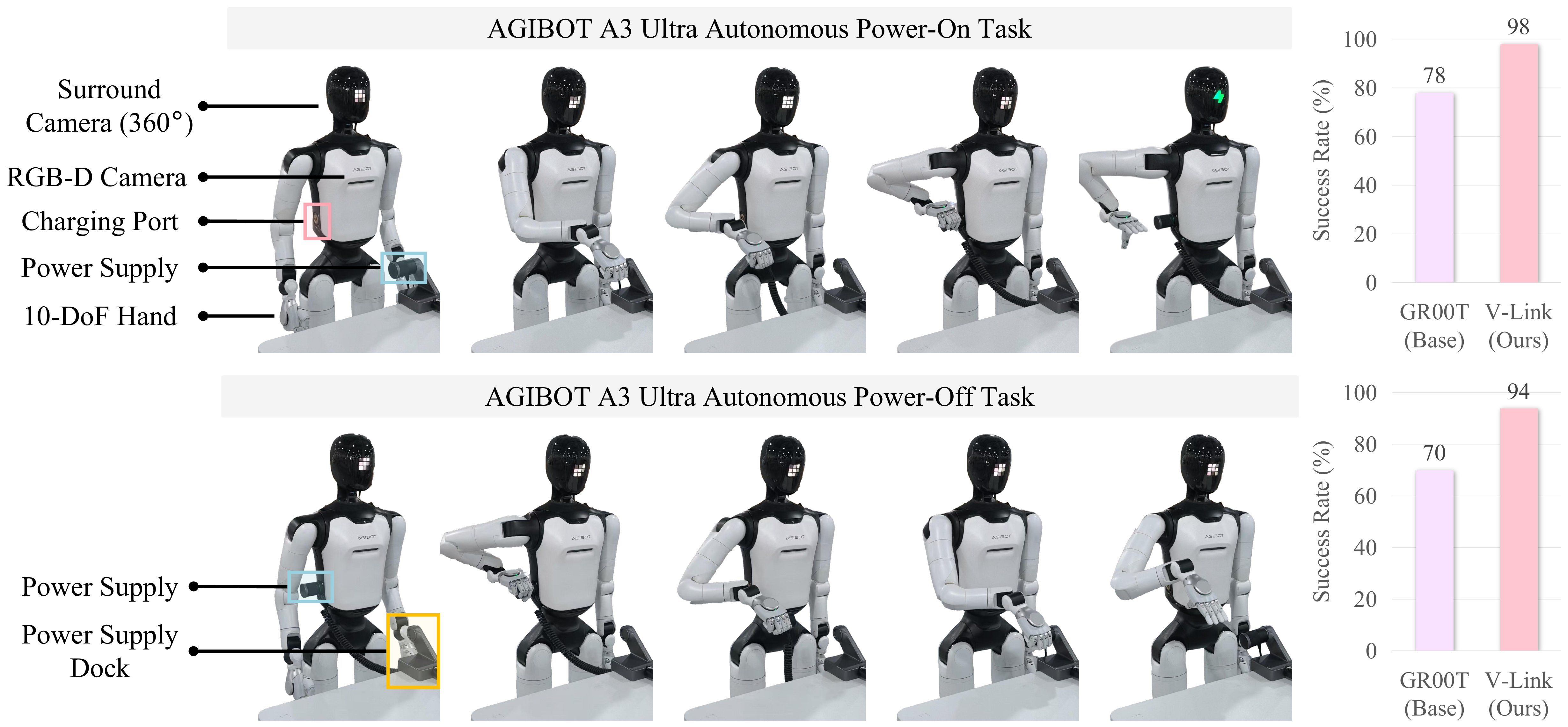}
    \caption{\textbf{Real-robot evaluation on the AGIBOT A3 Ultra.} Over 50 consecutive trials per task, V-Link achieves 98\%/94\% success rate on autonomous power-on/off tasks, versus 78\%/70\% for base model GR00T N1.6.}
    \vspace{-0.2cm}
    \label{A3U}
\end{figure*}

\subsection{Real-Robot Experiments}
We evaluate V-Link on the AGIBOT A3 Ultra in 50 consecutive trials per task. For training, Lingbot-Depth \cite{lingbot-depth2026} completes missing RGB-D depth caused by sensor dropouts and reflective surfaces, while GroundingDINO and SAM3 generate key-object segmentation pseudo-GT, with other pixels labeled as background. Without auxiliary heads or pseudo-GT at inference, V-Link achieves 98\%/94\% success on power-on/off, outperforming GR00T N1.6 at 78\%/70\%.
\section{Conclusion}
\label{sec:conclusion}
This work reveals a previously overlooked visual representation accessibility bottleneck in dual-system VLAs: VL-to-A transfer limits Action DiT’s access to 3D geometric and 2D semantic information encoded in VLM features. To address this issue, we propose V-Link, which learns complementary Spatial and Semantic Query representations and injects them asymmetrically into Action DiT, without requiring auxiliary heads or annotations at inference. Experiments across three simulation benchmarks and real-world humanoid manipulation demonstrate consistent gains with minimal inference overhead. These results establish representation recovery at the VL-to-A interface as an effective approach for precise and efficient robot manipulation.



\bibliographystyle{IEEEtran}
\bibliography{IEEEabrv}

@article{UniVLA,
  title={Unified vision-language-action model},
  author={Wang, Yuqi and Li, Xinghang and Wang, Wenxuan and Zhang, Junbo and Li, Yingyan and Chen, Yuntao and Wang, Xinlong and Zhang, Zhaoxiang},
  journal={arXiv:2506.19850},
  year={2025}
}

@article{Cosmos_policy,
  title={Cosmos policy: Fine-tuning video models for visuomotor control and planning},
  author={Kim, Moo Jin and Gao, Yihuai and Lin, Tsung-Yi and Lin, Yen-Chen and Ge, Yunhao and Lam, Grace and Liang, Percy and Song, Shuran and Liu, Ming-Yu and Finn, Chelsea and others},
  journal={arXiv:2601.16163},
  year={2026}
}

@article{pi_05,
  title={$\pi_{0.5}$: a Vision-Language-Action Model with Open-World Generalization},
  author={Black, Kevin and Brown, Noah and Darpinian, James and Dhabalia, Karan and Driess, Danny and Esmail, Adnan and Equi, Michael and Finn, Chelsea and Fusai, Niccolo and others},
  journal={arXiv:2504.16054},
  year={2025}
}

@article{LingBot-VLA,
  title={A Pragmatic VLA Foundation Model},
  author={Wu, Wei and Lu, Fan and Wang, Yunnan and Yang, Shuai and Liu, Shi and Wang, Fangjing and Zhu, Qian and Sun, He and Wang, Yong and Ma, Shuailei and others},
  journal={arXiv:2601.18692},
  year={2026}
}

@article{Gr00t_n1,
  title={Gr00t n1: An open foundation model for generalist humanoid robots},
  author={Bjorck, Johan and Casta{\~n}eda, Fernando and Cherniadev, Nikita and Da, Xingye and Ding, Runyu and Fan, Linxi and Fang, Yu and Fox, Dieter and Hu, Fengyuan and Huang, Spencer and others},
  journal={arXiv:2503.14734},
  year={2025}
}

@article{pi0_Fast,
  title={Fast: Efficient action tokenization for vision-language-action models},
  author={Pertsch, Karl and Stachowicz, Kyle and Ichter, Brian and Driess, Danny and Nair, Suraj and Vuong, Quan and Mees, Oier and Finn, Chelsea and Levine, Sergey},
  journal={arXiv:2501.09747},
  year={2025}
}

@article{Abot-m0,
  title={Abot-m0: Vla foundation model for robotic manipulation with action manifold learning},
  author={Yang, Yandan and Zeng, Shuang and Lin, Tong and Chang, Xinyuan and Qi, Dekang and Xiao, Junjin and Liu, Haoyun and Chen, Ronghan and Chen, Yuzhi and Huo, Dongjie and others},
  journal={arXiv:2602.11236},
  year={2026}
}

@article{Vla-cache,
  title={Vla-cache: Efficient vision-language-action manipulation via adaptive token caching},
  author={Xu, Siyu and Wang, Yunke and Xia, Chenghao and Zhu, Dihao and Huang, Tao and Xu, Chang},
  journal={Advances in Neural Information Processing Systems},
  volume={38},
  pages={164448-164473},
  year={2026}
}

@article{Sp-vla,
  title={Sp-vla: A joint model scheduling and token pruning approach for vla model acceleration},
  author={Li, Ye and Meng, Yuan and Sun, Zewen and Ji, Kangye and Tang, Chen and Fan, Jiajun and Ma, Xinzhu and Xia, Shutao and Wang, Zhi and Zhu, Wenwu},
  journal={arXiv:2506.12723},
  year={2025}
}

@article{UniFS,
  title={UniFS: Unified Fast-to-Slow Hierarchical Architecture for Vision-Language-Action Models},
  author={Sun, Lin and Guan, Zhiwei and Wang, Conglin and Chen, Zihong and Yu, Jianhai and Li, Zongsheng and He, Boyong and Sun, Tao and Cao, Jiale and Liu, Lige},
  journal={arXiv:2606.22794},
  year={2026}
}

@article{GEAR-VLA,
  title={GEAR-VLA: Learning Geometry-Aware Action Representations for Generalizable Robotic Manipulation},
  author={Zhang, Yuan and Zhang, Shiqi and Shen, Yedong and Dong, Shuai and Deng, Jiajun and Zhang, Xin and Gao, Yuxuan and Wu, Jiajia and Nie, Xin and Cheng, Zhiyuan and others},
  journal={arXiv:2606.08530},
  year={2026}
}

@article{Spatialvla,
  title={Spatialvla: Exploring spatial representations for visual-language-action model},
  author={Qu, Delin and Song, Haoming and Chen, Qizhi and Yao, Yuanqi and Ye, Xinyi and Ding, Yan and Wang, Zhigang and Gu, JiaYuan and Zhao, Bin and Wang, Dong and others},
  journal={arXiv:2501.15830},
  year={2025}
}

@article{Evo-0,
  title={Evo-0: Vision-language-action model with implicit spatial understanding},
  author={Lin, Tao and Li, Gen and Zhong, Yilei and Zou, Yanwen and Du, Yuxin and Liu, Jiting and Gu, Encheng and Zhao, Bo},
  journal={arXiv:2507.00416},
  year={2025}
}

@article{Spatial-forcing,
  title={Spatial forcing: Implicit spatial representation alignment for vision-language-action model},
  author={Li, Fuhao and Song, Wenxuan and Zhao, Han and Wang, Jingbo and Ding, Pengxiang and Wang, Donglin and Zeng, Long and Li, Haoang},
  journal={arXiv:2510.12276},
  year={2025}
}

@inproceedings{Vggt,
  title={Vggt: Visual geometry grounded transformer},
  author={Wang, Jianyuan and Chen, Minghao and Karaev, Nikita and Vedaldi, Andrea and Rupprecht, Christian and Novotny, David},
  booktitle={Proceedings of the Computer Vision and Pattern Recognition Conference},
  pages={5294-5306},
  year={2025}
}

@article{Pointvla,
  title={Pointvla: Injecting the 3d world into vision-language-action models},
  author={Li, Chengmeng and Wen, Junjie and Peng, Yaxin and Peng, Yan and Zhu, Yichen},
  journal={IEEE Robotics and Automation Letters},
  volume={11},
  number={3},
  pages={2506-2513},
  year={2026}
}

@inproceedings{Vla-adapter,
  title={Vla-adapter: An effective paradigm for tiny-scale vision-language-action model},
  author={Wang, Yihao and Ding, Pengxiang and Li, Lingxiao and Cui, Can and Ge, Zirui and Tong, Xinyang and Song, Wenxuan and Zhao, Han and Zhao, Wei and Hou, Pengxu and others},
  booktitle={Proceedings of the AAAI conference on artificial intelligence},
  volume={40},
  number={22},
  pages={18638-18646},
  year={2026}
}

@article{Libero,
  title={Libero: Benchmarking knowledge transfer for lifelong robot learning},
  author={Liu, Bo and Zhu, Yifeng and Gao, Chongkai and Feng, Yihao and Liu, Qiang and Zhu, Yuke and Stone, Peter},
  journal={Advances in Neural Information Processing Systems},
  volume={36},
  pages={44776-44791},
  year={2023}
}

@article{Libero-plus,
  title={Libero-plus: In-depth robustness analysis of vision-language-action models},
  author={Fei, Senyu and Wang, Siyin and Shi, Junhao and Dai, Zihao and Cai, Jikun and Qian, Pengfang and Ji, Li and He, Xinzhe and Zhang, Shiduo and Fei, Zhaoye and others},
  journal={arXiv:2510.13626},
  year={2025}
}

@article{Robotwin-2.0,
  title={Robotwin 2.0: A scalable data generator and benchmark with strong domain randomization for robust bimanual robotic manipulation},
  author={Chen, Tianxing and Chen, Zanxin and Chen, Baijun and Cai, Zijian and Liu, Yibin and Li, Zixuan and Liang, Qiwei and Lin, Xianliang and Ge, Yiheng and Gu, Zhenyu and others},
  journal={arXiv:2506.18088},
  year={2025}
}

@article{lingbot-depth2026,
  title={Masked Depth Modeling for Spatial Perception},
  author={Tan, Bin and Sun, Changjiang and Qin, Xiage and Adai, Hanat and Fu, Zelin and Zhou, Tianxiang and Zhang, Han and Xu, Yinghao and Zhu, Xing and Shen, Yujun and Xue, Nan},
  journal={arXiv:2601.17895},
  year={2026}
}

@article{Openvla,
  title={Openvla: An open-source vision-language-action model},
  author={Kim, Moo Jin and Pertsch, Karl and Karamcheti, Siddharth and Xiao, Ted and Balakrishna, Ashwin and Nair, Suraj and Rafailov, Rafael and Foster, Ethan and Lam, Grace and Sanketi, Pannag and others},
  journal={arXiv:2406.09246},
  year={2024}
}

@inproceedings{Dita,
  title={Dita: Scaling diffusion transformer for generalist vision-language-action policy},
  author={Hou, Zhi and Zhang, Tianyi and Xiong, Yuwen and Duan, Haonan and Pu, Hengjun and Tong, Ronglei and Zhao, Chengyang and Zhu, Xizhou and Qiao, Yu and Dai, Jifeng and others},
  booktitle={2025 IEEE/CVF International Conference on Computer Vision (ICCV)},
  pages={7686-7697},
  year={2025}
}

@inproceedings{Cot-vla,
  title={Cot-vla: Visual chain-of-thought reasoning for vision-language-action models},
  author={Zhao, Qingqing and Lu, Yao and Kim, Moo Jin and Fu, Zipeng and Zhang, Zhuoyang and Wu, Yecheng and Li, Zhaoshuo and Ma, Qianli and Han, Song and Finn, Chelsea and others},
  booktitle={2025 IEEE/CVF Conference on Computer Vision and Pattern Recognition (CVPR)},
  pages={1702-1713},
  year={2025}
}

@article{pi_0,
  title={$\pi_0$: A Vision-Language-Action Flow Model for General Robot Control},
  author={Black, Kevin and Brown, Noah and Driess, Danny and Esmail, Adnan and Equi, Michael and Finn, Chelsea and Fusai, Niccolo and Groom, Lachy and Hausman, Karol and Ichter, Brian and others},
  journal={arXiv:2410.24164},
  year={2024}
}

@article{OpenVLA-OFT,
  title={Fine-tuning vision-language-action models: Optimizing speed and success},
  author={Kim, Moo Jin and Finn, Chelsea and Liang, Percy},
  journal={arXiv:2502.19645},
  year={2025}
}

@article{Geovla,
  title={Geovla: Empowering 3d representations in vision-language-action models},
  author={Sun, Lin and Xie, Bin and Liu, Yingfei and Shi, Hao and Wang, Tiancai and Cao, Jiale},
  journal={arXiv:2508.09071},
  year={2025}
}

@article{3d-cavla,
  title={3d cavla: Leveraging depth and 3d context to generalize vision language action models for unseen tasks},
  author={Bhat, Vineet and Lan, Yu-Hsiang and Krishnamurthy, Prashanth and Karri, Ramesh and Khorrami, Farshad},
  journal={arXiv:2505.05800},
  year={2025}
}

@article{Evo-depth,
  title={Evo-depth: A lightweight depth-enhanced vision-language-action model},
  author={Lin, Tao and Du, Yuxin and Liu, Jiting and Zhu, Nuobei and Li, Yunhe and Fu, Yuqian and Chen, Yinxinyu and Cai, Hongyi and Ye, Zewei and Cheng, Bing and others},
  journal={arXiv:2605.14950},
  year={2026}
}

@article{Xiaomi-robotics-0,
  title={Xiaomi-robotics-0: An open-sourced vision-language-action model with real-time execution},
  author={Cai, Rui and Guo, Jun and He, Xinze and Jin, Piaopiao and Li, Jie and Lin, Bingxuan and Liu, Futeng and Liu, Wei and Ma, Fei and Ma, Kun and others},
  journal={arXiv:2602.12684},
  year={2026}
}

@article{EventVLA,
  title={EventVLA: Event-Driven Visual Evidence Memory for Long-Horizon Vision-Language-Action Policies},
  author={Yang, Ganlin and Tu, Zhangzheng and Yang, Yuqiang and Mao, Sitong and Dong, Junyi and Chen, Tianxing and Peng, Jiaqi and Xiong, Jing and Cao, Jiafei and Dai, Jifeng and others},
  journal={arXiv:2606.20092},
  year={2026}
}

\end{document}